# A Novel Approach of Harris Corner Detection of Noisy Images using Adaptive Wavelet Thresholding Technique

[1]Nilanjan Dey, [2]Pradipti Nandi, [3]Nilanjana Barman

[1]Dept. of IT, JIS College of Engineering, Kalyani, West Bengal, India
[2,3]Dept. of CSE, JIS College of Engineering, Kalyani, West Bengal, India

## Abstract
In this paper we propose a method of corner detection for obtaining features which is required to track and recognize objects within a noisy image. Corner detection of noisy images is a challenging task in image processing. Natural images often get corrupted by noise during acquisition and transmission. Though Corner detection of these noisy images does not provide desired results, hence de-noising is required. Adaptive wavelet thresholding approach is applied for the same.

## Keywords
Wavelet, De-noising, Harris Corner Detection, Bayes Soft threshold

## I. Introduction
A corner is a point for which there are two dominant and different edge directions in the vicinity of the point. In simpler terms, a corner can be defined as the intersection of two edges, where an edge is a sharp change in image brightness. Generally termed as interest point detection, corner detection is a methodology used within computer vision systems to obtain certain kinds of features from a given image. The initial operator concept of "points of interest" in an image, which could be used to locate matching regions in different images, was developed by Hans P. Moravec in 1977. The Moravec operator is considered to be a corner detector because it defines interest points as points where there are large intensity variations in all directions. For a human, it is easier to identify a "corner", but a mathematical detection is required in case of algorithms. Chris Harris and Mike Stephens in 1988 improved upon Moravec's corner detector by taking into account the differential of the corner score with respect to direction directly, instead of using shifted patches. Moravec only considered shifts in discrete 45 degree angles whereas Harris considered all directions. Harris detector has proved to be more accurate in distinguishing between edges and corners. He used a circular Gaussian window to reduce noise. Still in cases of noisy Images, it's difficult to find out the exact number of corners. One of the most conventional ways of image de-noising is using linear filters like Wiener filter. In the presence of additive noise the resultant noisy image, through linear filters, gets blurred and smoothed with poor feature localization and incomplete noise suppression. To overcome these limitations, nonlinear filters have been proposed like adaptive wavelet thresholding approach. Adaptive wavelet thresholding approach gives a very good result for the same. Wavelet Transformation has its own excellent space-frequency localization property and thresholding removes coefficients that are inconsiderably relative to some adaptive data-driven threshold.

## II. Discrete Wavelet Transformation
The wavelet transform describes a multi-resolution decomposition process in terms of expansion of an image onto a set of wavelet basis functions. Discrete Wavelet Transformation has its own excellent space frequency localization property. Applying DWT in 2D images corresponds to 2D filter image processing in each dimension. The input image is divided into 4 non-overlapping multi-resolution sub-bands by the filters, namely LL1 (Approximation coefficients), LH1 (vertical details), HL1 (horizontal details) and HH1 (diagonal details). The sub-band (LL1) is processed further to obtain the next coarser scale of wavelet coefficients, until some final scale "N" is reached. When "N" is reached, we'll have 3N+1 sub-bands consisting of the multi-resolution sub-bands (LLN) and (LHX), (HLX) and (HHX) where "X" ranges from 1 until "N". Generally most of the Image energy is stored in these sub-bands.

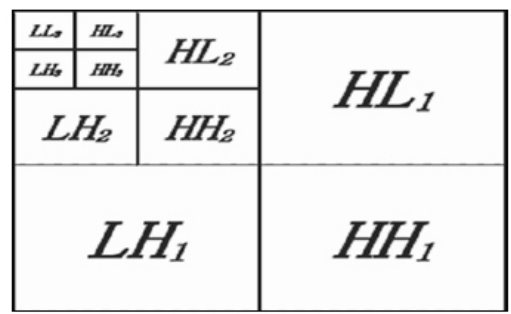

Fig. 1: Three phase decomposition using DWT

The Haar wavelet is also the simplest possible wavelet. Haar wavelet is not continuous, and therefore not differentiable. This property can, however, be an advantage for the analysis of signals with sudden transitions.

## III. Wavelet Thresholding
The concept of wavelet de-noising technique can be given as follows. Assuming that the noisy data is given by the following equation,
X (t) = S (t) + N (t)
Where, S (t) is the uncorrupted signal with additive noise N (t).
Let W (.) and W-1(.) denote the forward and inverse wavelet transform operators.
Let D (., $\lambda$) denote the de-noising operator with threshold $\lambda$. We intend to de-noise X (t) to recover Ŝ (t) as an estimate of S (t).
The technique can be summarized in three steps
Y = W(X)　　　　　　　　　　　　　　　　　　(2)
Z = D(Y, $\lambda$)　　　　　　　　　　　　　　　　　(3)
Ŝ = W-1 (Z)　　　　　　　　　　　　　　　　　　(4)
D (., $\lambda$) being the thresholding operator and $\lambda$ being the threshold.
A signal estimation technique that exploits the potential of wavelet transform required for signal de-noising is called Wavelet Thresholding (1, 2, 3). It de-noises by eradicating coefficients that are extraneous relative to some threshold. There are two types of recurrently used thresholding methods, namely hard and soft thresholding [4, 5].

The Hard thresholding method zeros the coefficients that are smaller than the threshold and leaves the other ones unchanged. On the other hand soft thresholding scales the remaining coefficients in order to form a continuous distribution of the coefficients centered on zero.
The hard thresholding operator is defined as
$$D(U, \lambda) = U \text{ for all } |U| > \lambda$$
Hard threshold is a keep or kill procedure and is more intuitively appealing. The hard-thresholding function chooses all wavelet coefficients that are greater than the given λ (threshold) and sets the other to zero. λ is chosen according to the signal energy and the noise variance ($\sigma^2$)

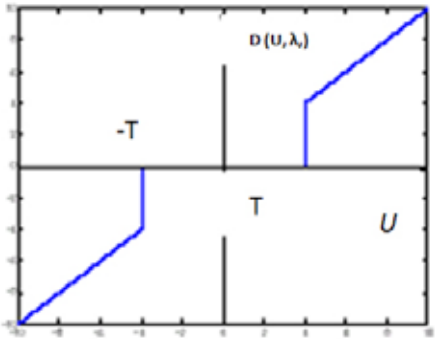

Fig. 2: Hard Thresholding

The soft thresholding operator is defined as
$$D(U, \lambda) = \text{sgn}(U) \max(0, |U| - \lambda)$$
Soft thresholding shrinks wavelets coefficients by λ towards zero.

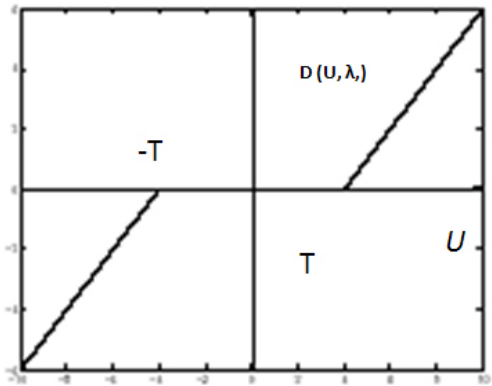

Fig. 3: Soft Thresholding

### IV. Bayes Shrink (BS)
Bayes Shrink, [6, 7] proposed by Chang Yu and Vetterli, is an adaptive data-driven threshold for image de-noising via wavelet soft-thresholding. Generalized Gaussian distribution (GGD) for the wavelet coefficients is assumed in each detail sub band. It is then tried to estimate the threshold T which minimizes the Bayesian Risk, which gives the name Bayes Shrink.
It uses soft thresholding which is done at each band of resolution in the wavelet decomposition. The Bayes threshold, $T_B$, is defined as
$$T_B = \sigma^2 / \sigma_s \quad (5)$$
Where $\sigma^2$ is the noise variance and $\sigma_s^2$ is the signal variance without noise. The noise Variance $\sigma^2$ is estimated from the sub band $HH_1$ by the median estimator

$$\hat{\sigma} = \frac{\text{median}(\{|g_{j-1,k}|: k = 0, 1, \ldots, 2^{j-1}-1\})}{0.6745} \quad (6)$$

where $g_{j-1,k}$ corresponds to the detail coefficients in the wavelet transform. From the definition of additive noise we have

$$w(x, y) = s(x, y) + n(x, y)$$
Since the noise and the signal are independent of each other, it can be stated that
$$\sigma_w^2 = \sigma_s^2 + \sigma^2$$

$\sigma_w^2$ can be computed as shown :

$$\sigma^2{}_w = \frac{1}{n^2} \sum_{x,y=1}^{n} w^2(x, y)$$

The variance of the signal, $\sigma_s^2$ is computed as

$$\sigma_s = \sqrt{\max(\sigma^2{}_w - \sigma^2, 0)}.$$

With $\sigma^2$ and $\sigma_s^2$, the Bayes threshold is computed from Equation (5).

### V. Harris Corner Detection
Harris corner detector [8, 9] is based on the local auto-correlation function of a signal which measures the local changes of the signal with patches shifted by a small amount in different directions. Given a shift (x, y) and a point the auto-correlation function is defined as

$$c(x,y) = \sum_W [I(x_i, y_i) - I(x_i + \Delta x, y_i + \Delta y)]^2 \quad (7)$$

Where $I(x_i, y_i)$ represent the image function and $(x_i, y_i)$ are the points in the window W centered on (x, y).
The shifted image is approximated by a Taylor expansion truncated to the first order terms

$$I(x_i + \Delta x, y_i + \Delta y) \approx [I(x_i, y_i) + [I_x(x_i, y_i) \ I_y(x_i, y_i)]] \begin{bmatrix} \Delta x \\ \Delta y \end{bmatrix} \quad (8)$$

where $I_x(x_i, y_i)$ and $I_y(x_i, y_i)$ indicate the partial derivatives in x and y respectively. With a filter like [-1, 0, 1] and [-1, 0, 1]$^T$, the partial derivates can be calculated from the image by substituting (8) in (7).

$$c(x,y) = [\Delta x \ \Delta y] \begin{bmatrix} \sum_W (I_x(x_i,y_i))^2 & \sum_W I_x(x_i,y_i) I_y(x_i,y_i) \\ \sum_W I_x(x_i,y_i) I_y(x_i,y_i) & \sum_W (I_y(x_i,y_i))^2 \end{bmatrix} \begin{bmatrix} \Delta x \\ \Delta y \end{bmatrix} = [\Delta x \ \Delta y] C(x,y) \begin{bmatrix} \Delta x \\ \Delta y \end{bmatrix}$$

C(x, y) the auto-correlation matrix which captures the intensity structure of the local neighborhood.
Let $\alpha_1$ and $\alpha_2$ be the eigenvalues of C(x, y), then we have 3 cases to consider:
1. Both eigenvalues are small means uniform region (constant intensity).
2. Both eigenvalues are high means Interest point (corner)
3. One eigen value is high means contour(edge)

To find out the interest points, Characterize corner response H(x, y) by Eigen values of C(x, y).
(i). C(x, y) is symmetric and positive definite that is $\alpha_1$ and $\alpha_2$ are >0
(ii). $\alpha_1 \alpha_2$ = det (C(x, y)) = AC –$B^2$
- $\alpha_1 + \alpha_2$ = trace(C(x, y)) = A + C
(iii). Harris suggested: That the
$$H_{cornerResponse} = \alpha_1 \alpha_2 - 0.04(\alpha_1 + \alpha_2)^2$$
Finally, it is needed to find out corner points as local maxima of the corner response.

## VI. Proposed Method

1. Perform 2-level Multi-wavelet decomposition of the image corrupted by Gaussian noise.
2. Apply Bayes Soft thresholding to the noisy coefficients.
3. Apply Harris corner detection on the de-noised image.

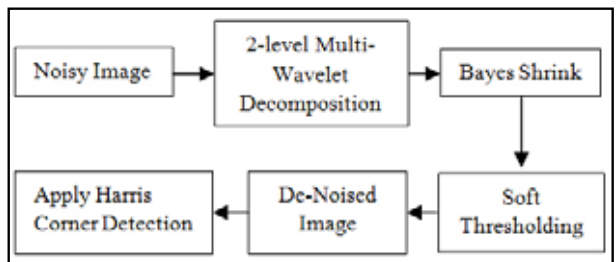

Fig. 4: Corner detection of De-noised image

## VII. Result and Discussions

Signal-to-noise ratio can be defined in a different manner in image processing where the numerator is the square of the peak value of the signal and the denominator equals the noise variance. Two of the error metrics used to compare the various image de-noising techniques is the Mean Square Error (MSE) and the Peak Signal to Noise Ratio (PSNR).

### A. Mean Square Error (MSE)

Mean Square Error is the measurement of average of the square of errors and is the cumulative squared error between the noisy and the original image.

$$MSE = \frac{1}{M \times N} \sum_{i=0}^{M-1} \sum_{j=0}^{N-1} (f(i,j) - g(i,j))^2$$

### B. Peak Signal to Noise Ratio (PSNR)

PSNR is a measure of the peak error. Peak Signal to Noise Ratio is the ratio of the square of the peak value the signal could have to the noise variance.

$$PSNR = 10 * \log_{10}\left(\frac{255^2}{MSE}\right)$$

A higher value of PSNR is good because of the superiority of the signal to that of the noise.

MSE and PSNR values of an image are evaluated after adding Gaussian and Speckle noise. The following tabulation shows the comparative study based on Wavelet thresholding techniques of different decomposition levels.

Table 1:

| Noise Type | Wavelet | Thres-holding | Level of Decom-position | PSNR |
|---|---|---|---|---|
| Gaussian | Haar | Bayes Soft | 1 | 22.8685 |
| | | | 2 | 23.6533 |
| Speckle | | | 1 | 23.0867 |
| | | | 2 | 23.6360 |
| Salt & Paper | | | 1 | 19.5202 |
| | | | 2 | 19.5405 |

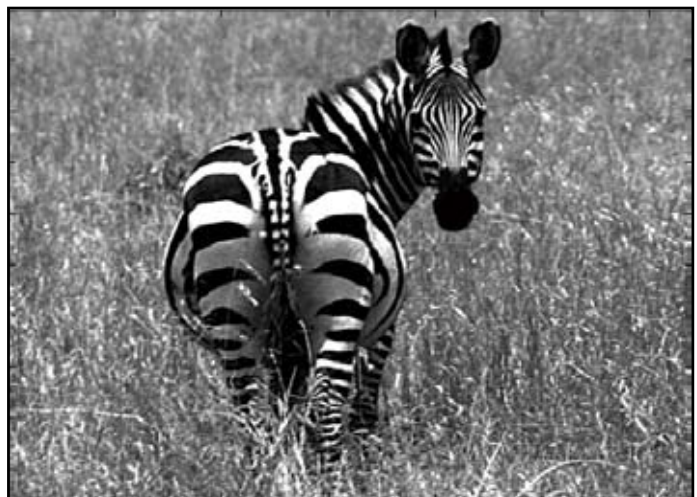
(a)

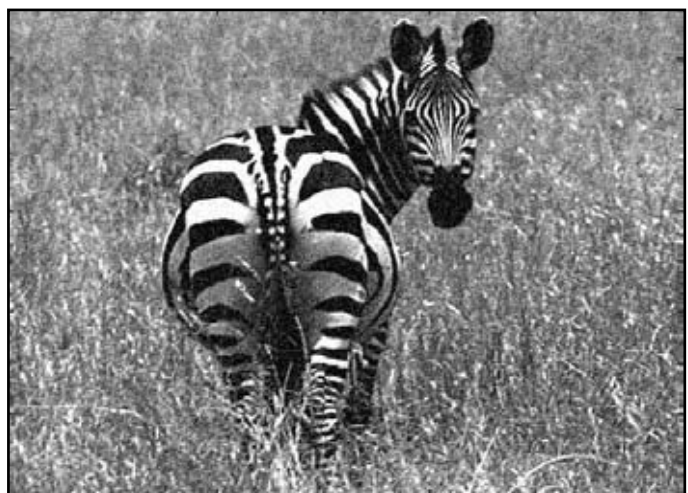
(b)

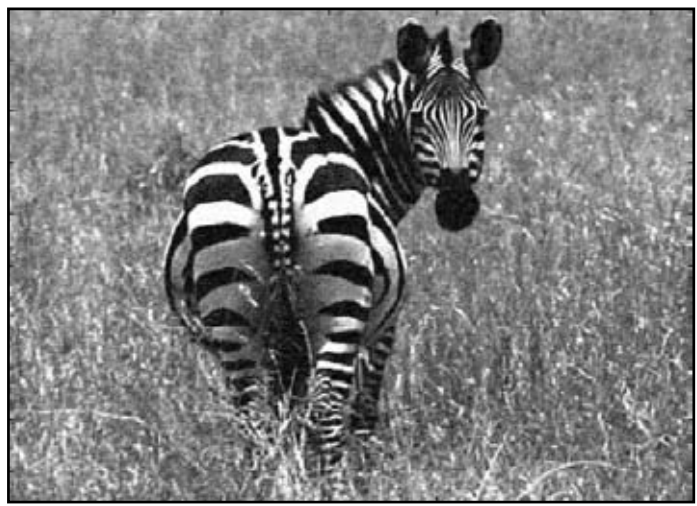
(c)

(a). Original Image (b). Noisy image (Gaussian) (c). Second level DWT decomposed and Bayes soft threshold noisy image
Fig. 5: Denoising method

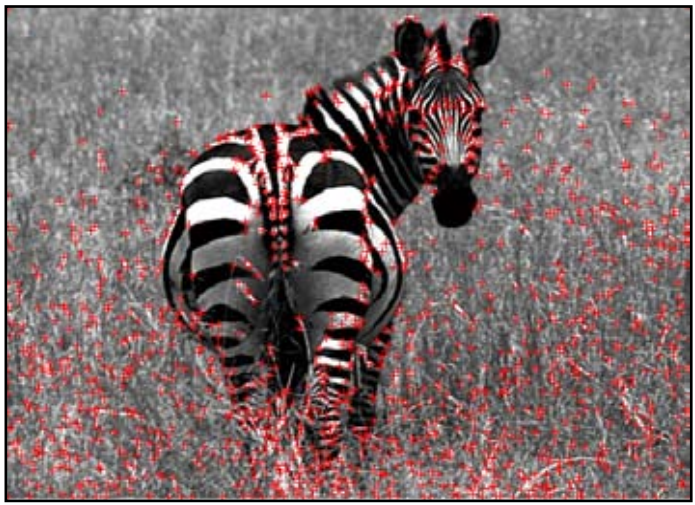

(a)

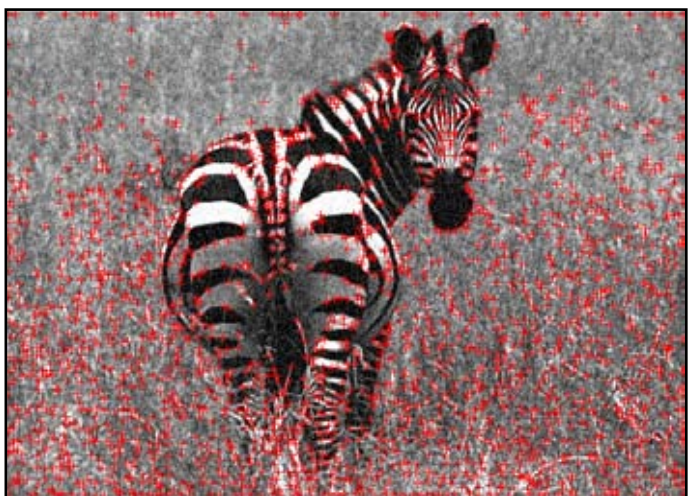

(b)

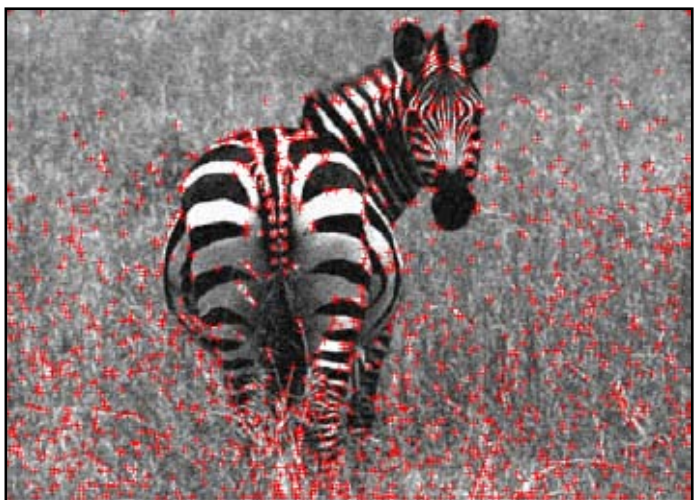

(c)

(a). Harris Corner Detection on original image (b). Harris Corner Detection on noise image (c). Harris Corner Detection on de-noised image using Bayes soft threshold method.
Fig. 6: Harris Corner Detection

Table 2:

| Image Type | | Harris Corner detected |
|---|---|---|
| Original | | 1329 |
| Noise | Gaussian (zero mean noise with 0.01 variance) | 2259 |
| | Speckle (mean 0 and variance v. The default for v is 0.04) | 2137 |
| | Salt & Pepper (0.05 noise density) | 2778 |

Table 3:

| Noise Type | Wavelet | Thres-holding | Level of Decom-position | Harris Corner detected |
|---|---|---|---|---|
| Gaussian | Haar | Bayes Soft | 1 | 1697 |
| | | | 2 | 1379 |
| Speckle | | | 1 | 1910 |
| | | | 2 | 1623 |
| Salt & Pepper | | | 1 | 2624 |
| | | | 2 | 2554 |

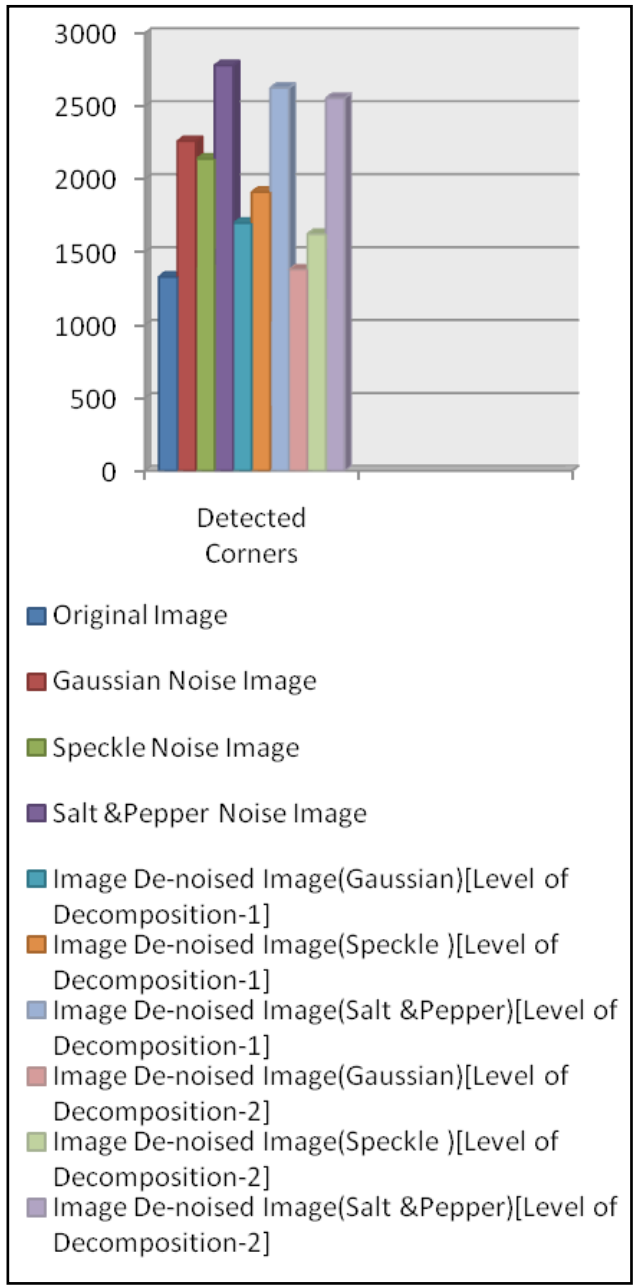

## VII. Conclusion

The BS method is effective for images including Gaussian noise. As the experimental result shows that the number of corner detected for obtaining features from the original image is near equal to the same with the number of points detected by de-noised image using BS method.

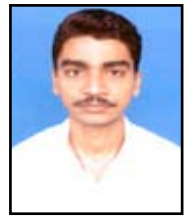

Nilanjan Dey is an Asst. Professor at Department of Information Technology, JIS College of Engineering, Kalyani, under West Bengal University of Technology, India. He holds an M.Tech degree and a B.Tech degree in Information Technology from West Bengal University of Technology, India. The Author has 3 years of teaching experience along with 1.2 yrs of Industrial experience. His research interests are image processing, Artificial intelligence, data mining, wavelet and computation. He has a number of research papers published in National & International Journals on Image Processing & Analysis.

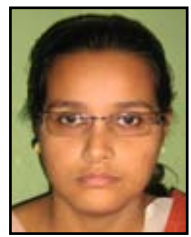

Pradipti Nandi a final year student of Computer Science and Engineering from JIS College of Engineering, Kalyani, under West Bengal University of Technology, India, currently appointed as the Programmer Analyst Trainee in Cognizant Technology Solutions. Pradipti is tremendously interested in field of image processing, network security, and hence tries to explore various possibilities in these fields to obtain improved results. This paper contributes as her first experience in publishing her work.

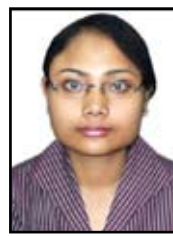

Nilanjana Barman a final year student of Computer Science and Engineering from JIS College of Engineering, Kalyani, under West Bengal University of Technology, India, currently appointed as the Programmer Analyst Trainee in Cognizant Technology Solutions. Her area of interest is image processing, digital signal processing & network security. This paper contributes as her first experience in publishing her work.